\pdfoutput=1

\documentclass[11pt]{article}

\usepackage{acl}

\usepackage{times}
\usepackage{latexsym}

\usepackage[T1]{fontenc}

\usepackage[utf8]{inputenc}

\usepackage{microtype}

\usepackage{inconsolata}

\usepackage{graphicx}
\usepackage{subcaption}

\usepackage[most]{tcolorbox}

\usepackage{booktabs}

\title{Does It Run and Is That \emph{Enough}? Revisiting Text-to-Chart Generation with a Multi-Agent Approach}

\author{James Ford \and Anthony Rios \\
  Department of Information Systems and Cyber Security \\
  The University of Texas at San Antonio  \\
  \texttt{\{james.ford, anthony.rios\}@utsa.edu}\\}

\begin{document}
\maketitle
\begin{abstract}
Large language models can translate natural-language chart descriptions into runnable code, yet approximately 15\% of the generated scripts still fail to execute, even after supervised fine-tuning and reinforcement learning. We investigate whether this persistent error rate stems from model limitations or from reliance on a single-prompt design. To explore this, we propose a lightweight multi-agent pipeline that separates drafting, execution, repair, and judgment, using only an off-the-shelf GPT-4o-mini model. On the \textsc{Text2Chart31} benchmark, our system reduces execution errors to 4.5\% within three repair iterations, outperforming the strongest fine-tuned baseline by nearly 5 percentage points while requiring significantly less compute. Similar performance is observed on the \textsc{ChartX} benchmark, with an error rate of 4.6\%, demonstrating strong generalization. Under current benchmarks, execution success appears largely solved. However, manual review reveals that 6 out of 100 sampled charts contain hallucinations, and an LLM-based accessibility audit shows that only 33.3\% (\textsc{Text2Chart31}) and 7.2\% (\textsc{ChartX}) of generated charts satisfy basic colorblindness guidelines. These findings suggest that future work should shift focus from execution reliability toward improving chart aesthetics, semantic fidelity, and accessibility.
\end{abstract}

\section{Introduction}


Natural language chart generation turns a request such as \emph{plot quarterly revenue by region} into runnable code and a complete figure.  A reliable solution would let non-programmers explore data, shorten the loops of professional analysts, and (potentially) create alternative renderings for low-vision users that are more accessible to screen readers.  Two conditions decide whether the community can call the task solved: the generated script must execute without error, and the visual output must match the description.

One public benchmark has guided much of the recent progress.  The dataset \textsc{Text2Chart31}~\cite{pesaran-zadeh-etal-2024-text2chart31} measures exact failure rates across thirty-one plot types along with various code similarity metrics. Zero-shot prompts for large language models (LLMs) initially failed on about forty percent of inputs. Supervised fine-tuning and reinforcement learning reduced that number below fifteen percent and brought pairwise plots close to six percent. Prompt engineering improved results a little more~\cite{zhang-etal-2024-gpt,koh2025c2scalableautofeedbackllmbased,10.1145/3664647.3680790,li-etal-2024-visualization}. Closed-source models such as GPT-4-turbo and Claude3Opus achieved similar levels, yet stubborn errors remained for complex surfaces and volumetric plots.  Work on chart captioning and visual question answering~\cite{ford-etal-2025-charting,YE202443,han2023chartllamamultimodalllmchart,kim2025simplotenhancingchartquestion,wang2024charxivchartinggapsrealistic} shows that even a small execution failure can break an entire downstream pipeline.

Instead of one large prompt, agent-based systems use several small prompts that plan, call tools, and verify interim results~\cite{ji2025socraticchartcooperatingmultiple,10753451}. Code-repair agents update faulty scripts after runtime errors and often succeed within a few steps~\cite{10.1109/ICSE48619.2023.00128,DEFITERODOMINGUEZ2024109291,bouzenia2024repairagentautonomousllmbasedagent,10440574}. This study asks two research questions.   First, whether chart generation, defined in its present benchmark setting, can be improved using a simple agent-based framework.  If the answer is yes, the field would need to revisit its evaluation criteria, shift resources away from training larger models for pure execution, and retire the current benchmarks in favour of more difficult scenarios such as noisy data or multi-step analytic sessions.

Second, if using traditional metrics and our approach effectively ``solves'' what objectives should guide new work when execution errors are already rare? For general image generation, could extend chart evaluation work to move beyond a binary ``runs or not'' view. Structural similarity (SSIM) scores reflect low-level alignment. At the same time, multimodal LLM judges rate perceptual fidelity and semantic match~\cite{10.1007/978-3-031-72970-6_15,10.1007/978-3-031-72904-1_9,10.1007/978-3-031-70533-5_26,10.1007/978-3-031-88720-8_22}. Furthermore, a focus on aesthetics, readability, and accessibility could produce charts that follow color-contrast guidelines for color-blind users, place legends to avoid overlap, and include alt-text or tactile representations for screen readers and embossing devices.  Such qualities matter for journalism, education, and government dashboards, where the chart must inform a wide audience rather than only run without crashing.

To answer these questions, we developed a lightweight multi-agent pipeline that uses an off-the-shelf GPT-4o-mini model without any additional training. A \textsc{Draft} agent generates Python/Matplotlib code from natural-language chart descriptions, and a \textsc{Repair} agent iteratively debugs and rewrites the code—up to three times—when execution fails. On both \textsc{Text2Chart31} and \textsc{ChartX}, the system reduces execution errors by nearly 5 absolute percentage points, outperforming fine-tuned baselines while preserving image quality as measured by SSIM and multimodal LLM judgment. A manual analysis of 100 sampled outputs found that 83\% were accurate, with most remaining issues related to minor stylistic differences rather than semantic or data errors. These findings suggest that agentic pipelines are not only more robust than single-shot prompting but also capable of producing high-quality, visually faithful charts.

Overall, we make the following contributions.
\begin{itemize}
    \item We introduce a multi-agent pipeline that achieves state-of-the-art execution success on two public benchmarks. Our system reduces the overall error rate of existing models by nearly 5\% with no drop in performance in visual quality.
    \item We provide empirical evidence that execution is largely ``solved'' in current benchmarks and perform a comprehensive analysis to explore the missing gaps in current evaluation that should be targeted in the future.
\end{itemize}

\section{Related Work}
\vspace{1mm} \noindent \textbf{Chart Generation and Evaluation.}
Large language models (LLMs) are increasingly used to automate chart generation from natural language, but major challenges persist in quality control and evaluation. Datasets such as \textsc{ChartQA}, \textsc{PlotQA}, \textsc{ChartLlama}, and \textsc{Charxiv}~\cite{masry-etal-2022-chartqa,Methani_2020_WACV,han2023chartllamamultimodalllmchart,wang2024charxivchartinggapsrealistic} establish benchmarks for chart reasoning, captioning, and data extraction. These corpora focus primarily on chart understanding and question answering. \citet{ford-etal-2025-charting} and \citet{YE202443} also examine chart captioning and visual QA.

To address text-to-chart generation directly, \textsc{Text2Chart31} and \textsc{ChartX} pair instructions with both code and image outputs, allowing complete evaluation pipelines~\cite{pesaran-zadeh-etal-2024-text2chart31,xia2025chartxchartvlmversatile}. Still, these datasets remain small and often require human references. Prompt engineering, CoT prompting~\cite{podo2024vrecslowcostllm4visrecommender,li2024visualizationgenerationlargelanguage}, and image-based analysis methods (e.g., SSIM) have been proposed to assess generated charts~\cite{10.1007/978-3-031-72970-6_15,10.1007/978-3-031-72904-1_9}. However, hallucinated or stylistically inconsistent outputs remain common~\cite{podo2024vievallmconceptualstack,10443572}. Moreover, recent work questions the reliability of multimodal LLM judges for chart evaluation~\cite{mukhopadhyay-etal-2024-unraveling,masry2024chartgemmavisualinstructiontuningchart}. This motivates the development of task-specific evaluation protocols combining low-level visual similarity (e.g., SSIM) as well as using LLM-as-a-judge~\cite{10.1007/978-3-031-70533-5_26,10.1007/978-3-031-88720-8_22}.

\vspace{1mm} \noindent \textbf{Code Generation and Repair.}
Chart generation with LLMs relies on producing syntactically and semantically valid code. Yet generated scripts often contain runtime errors, prompting the need for automated debugging frameworks. Several works propose feedback-driven or error-trace-guided repairs~\cite{10.1109/ICSE48619.2023.00128,DEFITERODOMINGUEZ2024109291,bouzenia2024repairagentautonomousllmbasedagent,10440574}. These methods fall into agentless, agentic, and retrieval-augmented categories~\cite{10968728}, but still suffer from semantic hallucination and misalignment with user intent~\cite{10734039}.


\vspace{1mm} \noindent \textbf{Agentic Chart Code Generation.}
Agentic systems introduce structured coordination among LLM components for planning, generation, and validation. Tool-augmented and multi-agent pipelines have been proposed across domains~\cite{cheng-etal-2024-small,shang-etal-2024-traveler,shen-etal-2024-small,zhang-etal-2024-omagent,zong-etal-2024-triad,10766492}. Open-source ecosystems such as \textsc{LangChain} and \textsc{CrewAI}~\cite{Langchain,CrewAI} provide infrastructure to implement agentic workflows with memory and tool use.

\begin{figure*}[t]
    \centering    \includegraphics[width=1\linewidth]{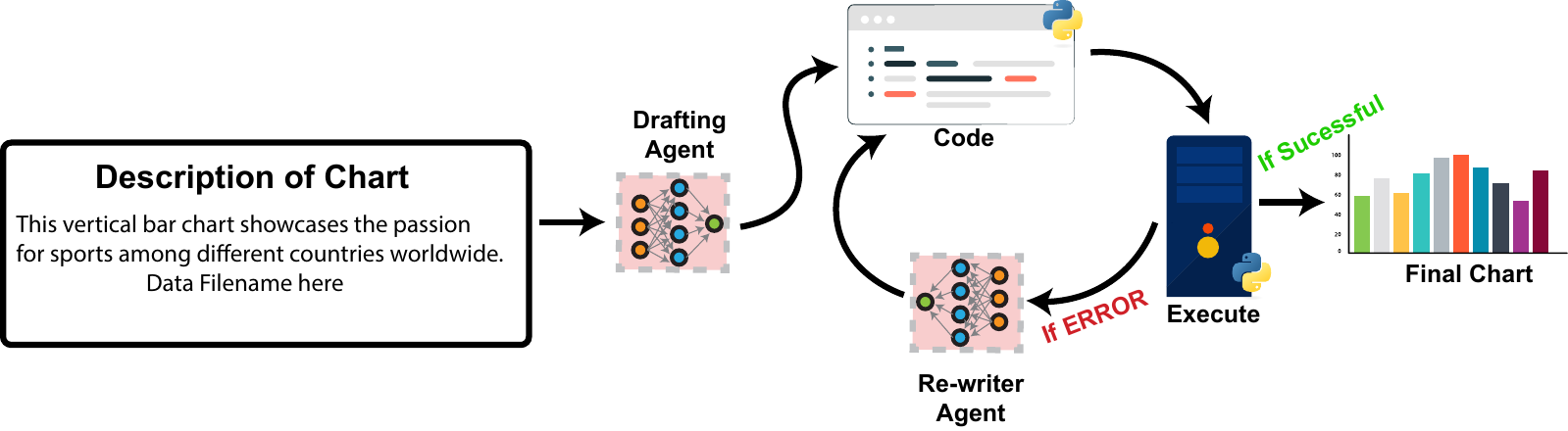}
\caption{\textbf{Multi-agent text-to-chart pipeline}.
A natural-language description is first passed to the \emph{Drafting Agent}, which synthesises runnable Python/Matplotlib code.
The script is executed; if a runtime exception occurs, the full traceback and source are forwarded to the \emph{Re-writer Agent}, which edits the code and resubmits it. This repair loop repeats (up to three iterations in our experiments) until the code runs successfully, producing the final chart with no human intervention.}    \label{fig:enter-label}\vspace{-1em}
\end{figure*}

In visualization tasks, \textsc{Socratic Chart} orchestrates chart question answering through multiple agents with access to visual and textual tools~\cite{ji2025socraticchartcooperatingmultiple}, while \textsc{LightVA} incorporates user feedback in an iterative refinement loop~\cite{10753451}. These systems exemplify how verification and repair agents can boost generation quality, even without direct reference data. Our work builds on this trend, highlighting the synergy between agentic reasoning, code correctness, and reference-free evaluation.

\begin{table}[t]
  \centering
  \resizebox{.8\linewidth}{!}{
  \begin{tabular}{lrrr}
    \toprule
    \textbf{Plot Type} & \textbf{Data Points} \\
    \midrule
    \multicolumn{2}{c}{\textbf{Text2Chart31}} \\ \midrule
    Pairwise Chart     & 472\\
    Statistical Distribution Chart     & 452\\
    Gridded Chart     & 192\\
    Irregularly Gridded Chart     & 148\\
    3D and Volumetric Chart     & 159  \\
    Total     & 1,423 \\
    \midrule
    \multicolumn{2}{c}{\textbf{ChartX}} \\ \midrule
    General Chart & 500 \\
    Fine-Grained Chart & 500 \\
    Specific Chart & 152 \\
    Total     & 1,152    \\\bottomrule
  \end{tabular}}
  \caption{Dataset statistics.}\vspace{-1em}
  \label{tab:stats}
\end{table}

\section{Data}
We use two benchmark datasets.  The statistics for the datasets used in our study are shown in Table~\ref{tab:stats}. 

\vspace{1mm} \noindent \textbf{Text2Chart31.} The Text2Chart31 dataset~\cite{pesaran-zadeh-etal-2024-text2chart31} consists of 11,128 data points with 31 types of plots, providing a robust environment to test the benefits of agentic visualizations.  The dataset contains text plot descriptions, Python code to create a charts, image files of the created charts, and (for 8,166 data points) csv data files as well as reasoning steps.  The dataset is segmented into 9,705 data points in the training set and 1,423 in the test set (for the purpose of this current study, only the test set was used). The dataset was created synthetically through the use of GPT-3.5-turbo and GPT-4.  Topics span nearly 50 scientific, political, economic, and popular culture subjects.  The 31 chart types are distributed in five categories, including pairwise, statistical, gridded, irregularly gridded, and 3D/volumetric data.

\vspace{1mm} \noindent \textbf{ChartX.} The second dataset is ChartX~\cite{xia2025chartxchartvlmversatile}. This benchmark data consists of 4,848 synthetically-generated chart examples in the validation set and 1,152 in the test set (similarly, for the purpose of this current study, only the test set was used), spanning 18 chart types. A total of 22 topics are represented in the charts, with similar cultural, economic, political, and scientific themes as the first dataset. The data file contains text descriptions of the charts, Python code, the raw data points, and chart image files.

\section{Methodology}

Figure 1 provides a high-level overview of our paper. Our framework has three main components. First, we use a \textsc{Drafting} agent that transforms the text into code to generate the chart. Second, the code is evaluated using a Python interpreter, and any resulting errors are passed to a \textsc{Re-Writer} agent; otherwise, if there is an error, the final chart is returned. If there was an error, the  \textsc{Re-Writer} agent will try to fix the error. This process is repeated until the code runs or a maximum number of iterations is hit. We describe each part in detail below.

\vspace{1mm} \noindent \textbf{Step 1: \textsc{Drafting} Agent.}
First, we generated baseline Python code from the provided text descriptions in the dataset using the closed-source GPT4o-mini LLM. We conducted both zero-shot and few-shot runs for the baseline.  In the zero-shot setting, the LLMs receive only the task instructions and the data without examples. The prompt follows the format from the Text2Chart31 paper~\cite{pesaran-zadeh-etal-2024-text2chart31}. Specifically, the system prompt instructs the model to generate Python code for data visualization as follows:
\begin{center}
\vspace{-.25em}
\tcbset{
    colframe=black,
    colback=white,
    boxrule=0.5mm,
    arc=3mm,
    width=1\linewidth,
    boxsep=5pt,
    left=5pt,
    right=5pt,
    top=5pt,
    bottom=5pt
}
\begin{tcolorbox}[colback=gray!5!white, colframe=black, fontupper=\small, width=\linewidth, boxsep=1pt, title=System Prompt]
\textit{You are good at generating complete python code from the given chart description.}
\end{tcolorbox}
\vspace{-.25em}
\end{center}
Next, the user instruction prompt then provides the specific chart details
\begin{center}
\vspace{-.25em}
\tcbset{
    colframe=black,
    colback=white,
    boxrule=0.5mm,
    arc=3mm,
    width=1\linewidth,
    boxsep=5pt,
    left=5pt,
    right=5pt,
    top=5pt,
    bottom=5pt
}
\begin{tcolorbox}[colback=gray!5!white, colframe=black, fontupper=\small, width=\linewidth, boxsep=1pt, title=User Instructions]
\textit{Your task is to generate a complete Python code for the given description. Make sure to include all necessary libraries.}

\texttt{Description: Description\_text}

\textit{Please generate the corresponding code that generates the plot that has the above description.}
\texttt{Code:}
\begin{verbatim}
```import matplotlib.pyplot as plt
import pandas as pd
import numpy as np
\end{verbatim}
\end{tcolorbox}
\vspace{-.25em}
\end{center}
where \texttt{Description\_text} is the provided text description which is passed to the model. An example of the text description is as follows: 
\begin{quote}
This vertical bar chart showcases the passion for sports among different countries worldwide. The chart represents the percentage of individuals in each country who identify themselves as avid sports fans. The data provides insights into the level of sports fanaticism across countries, offering a comparative analysis. 

The X-axis denotes the countries included in the dataset, labeled as 'Country', while the Y-axis signifies the percentage of individuals who consider themselves passionate sports fans, denoted as 'Percentage of Sports Fanatics'. 

The data for this graph is stored in a CSV file named 'sports\_fanatics.csv', which includes five columns: 'Country', 'Percentage of Sports Fanatics'. Here are the first six rows of the dataset: 

Country,Percentage of Sports Fanatics\\
United States,45\\
Germany,35\\
...
\end{quote}

This prompt directs the LLM to generate the appropriate Python code to create the chart per the specifications. The system prompt is mentioned once in the few-shot setting, and two in-context examples are provided. These examples are supplied in pairs, containing (1) the user instruction with the chart description and data file and (2) the code associated with the data files to generate the chart.

\begin{table*}[ht]
  \centering
  \resizebox{\textwidth}{!}{
  \begin{tabular}{lrrrrrrrrr}
    \toprule
    & \multicolumn{5}{c}{\textbf{Error Ratio}} & \multicolumn{2}{c}{\textbf{Code Similarity}} \\ \cmidrule(lr){2-6} \cmidrule(lr){7-9}
    \textbf{} & \textbf{} & \textbf{Statistical} & \textbf{(Irregularly)} & \textbf{3D and} \\    
    \textbf{Model} & \textbf{Pairwise} & \textbf{distribution} & \textbf{gridded} & \textbf{Volumetric} & \textbf{Total} & \textbf{METEOR} & \textbf{CodeBLEU} \\
    \midrule
    CLI-7B   &   22.67\    &  29.42\    & 77.94\  & 52.20\   & 41.32\  & 0.485\    & 0.402\  \\
    L3I-8B   & 20.76\    & 28.98\    &  66.76\  &  34.59\   & 35.91\    & 0.519\  &  0.437\  \\
    SFT: L3I-8B   & 19.07\    &  13.27\    &   13.53\  &  20.75\   & 16.09\    &  0.562\  &  0.464\  \\
    SFT+RLpref: L3I-8B   & 13.14\    &  11.50\    &  15.00\  &  26.42\   & 14.55\    &  0.567\  &  0.461\  \\ \midrule
    CLI-13B   &   18.86\    &  29.42\    & 71.76\  & 57.23\   &  39.14\  & 0.489\    &  0.413\  \\
    StarCoder-15.5B   &    23.31\    &  32.08\    & 51.18\  & 25.16\   &  32.89\  &  0.347\    &   0.328\  \\
    InstructCodeGen-16B   &    38.56\    &  45.13\    & 62.94\  & 40.25\   &  46.66\  &   0.388\    &    0.330\  \\
    SFT: CLI-13B   &     6.36\    &   6.19\    &  12.06\  & 22.64\   &  9.49\  &    \textbf{0.581}\    &     \textbf{0.481}\  \\
    SFT+RLpref: CLI-13B   &     6.36\    &   5.53\    &  12.35\  & 21.38\   &  9.21\  &     0.566\    &      0.467\  \\ \midrule
    GPT-3.5-turbo   &   11.02\    &  13.50\    & 28.82\  & 19.59\   &  18.62\  & 0.524\    &  0.453\  \\
    GPT-4-0613   &   13.56\    &  11.06\    & 28.53\  & 39.62\   &  19.26\  &  0.535\    &   0.441\  \\
    GPT-4-turbo   &   11.02\    &  14.16\    &  11.76\  &  29.56\   &   14.27\  &  0.540\    &    0.448\  \\
    GPT-4o   &   13.98\    &  6.86\    &   13.53\  &   26.42\   &   13.00\  &   0.552\    &    0.450\  \\
    Claude3Opus   &   7.84\    &  7.74\    &   30.59\  &    23.27\   &   14.90\  &    0.515\    &    0.435\  \\ \midrule
      ZS GPT 4o-mini Baseline &  13.35 & \   5.97 & \   18.53 & \ 35.85 & \ 14.76  & \  0.542  & \ 0.407 & \   \\ 
     ZS GPT 4o-mini Agentic &  2.54 & \   5.09 & \    9.41 & \  18.24 & \   6.75 & \   0.510 & \  0.406 & \   \\
     FS GPT 4o-mini Baseline & 14.62  & \  8.85  & \  12.45  & \ 31.45 & \ 14.13  & \  0.557  & \ 0.433 & \   \\ 
     FS GPT 4o-mini Agentic  & \textbf{1.48} & \  \textbf{3.76}  & \  \textbf{5.59}  & \ \textbf{13.21} & \  \textbf{4.50} & \  0.532 & \ 0.447 & \   \\ 
     \bottomrule
  \end{tabular}}
\caption{Performance of baseline, fine-tuned (SFT), and preference-optimized (SFT\,+\,RLpref) models on the \textsc{Text2Chart-31} benchmark. Columns 2–6 report error ratios (↓) for Pairwise, Statistical-Distribution, Irregularly-Gridded, 3D/Volumetric, and Overall errors; lower values indicate more accurate chart generation. Columns 7–8 give code-similarity scores (↑) using \textsc{METEOR} and \textsc{CodeBLEU}, where higher is better.  ``ZS'' denotes zero-shot models (all closed-source LLMs are ZS if not noted as FS), ``FS'' few-shot models, and ``Agentic'' indicates our agent-based prompting strategy. Best results for each metric are \textbf{bolded}.}\vspace{-1em}
  \label{tab:text2chart31-res}
\end{table*}

\vspace{1mm} \noindent \textbf{Step 2: \textsc{Re-writer} Agent.}
Second, the code generated from Step 1 is passed to a Python interpreter. If the code runs, the final chart is generated. However, if the code results in an error, the error and original code are passed to the \textsc{Re-writer} Agent. Specifically, for the instances where execution failed, the supervising agent sends the code and the error message to the re-writer agent to correct and re-execute the code. The process is repeated for a maximum of three iterations.   The system prompt is shown below (full prompt in the Appendix):
\begin{center}
\vspace{-.25em}
\tcbset{
    colframe=black,
    colback=white,
    boxrule=0.5mm,
    arc=3mm,
    width=1\linewidth,
    boxsep=5pt,
    left=5pt,
    right=5pt,
    top=5pt,
    bottom=5pt
}
\begin{tcolorbox}[colback=gray!5!white, colframe=black, fontupper=\small, width=\linewidth, boxsep=1pt, title=System Prompt]
        You are an expert Python code rewriter. Your task is to rewrite Python code based strictly on the user's suggestions.\\
        - DO NOT modify any part of the code that is not explicitly mentioned in the suggestion.\\
        - Ensure that the rewritten code is functional, error-free, and adheres to Python syntax rules (e.g., indentation, brackets, braces).\\
        - Return ONLY the complete revised code without explanations, comments, or Markdown formatting.\\
        - Follow instructions EXACTLY as provided.
\end{tcolorbox}
\vspace{-.25em}
\end{center}

\begin{table}[t]
  \centering
  \resizebox{.8\linewidth}{!}{
  \begin{tabular}{lrrr}
    \toprule
    \textbf{Model} & \textbf{Error Rate} \\
    \midrule
    ZS GPT 4o-mini Baseline & 11.11 \\
    ZS GPT 4o-mini Agentic & \textbf{3.13} \\
    FS GPT 4o-mini Baseline & 9.38 \\
    FS GPT 4o-mini Agentic & 4.60 \\\bottomrule
  \end{tabular}} 
\caption{Overall chart generation error rate (↓) on the \textsc{ChartX} benchmark. ``ZS'' denotes zero-shot models, ``FS'' few-shot models, and ``Agentic'' indicates our agent-based prompting strategy. Lower values mean fewer generation errors; the best score is \textbf{bolded}.}\vspace{-1em}
  \label{tab:chartx-res}
\end{table}

\section{Results}

\vspace{1mm} \noindent \textbf{Experimental Details.} All experiments are implemented using the  LangChain framework~\cite{Langchain}. Overall, our experiments cost $\sim$\$150, including all experiments that did and did not work.

\vspace{1mm} \noindent \textbf{Evaluation Metrics.} For our main metrics, we use code similarity and execution error rates. Execution error rates is calculated as the proportion of code snippets that do not run in the Python interpreter. Code similarity metrics are calculated with METEOR~\cite{banerjee-lavie-2005-meteor} and CodeBLEU~\cite{ren2020codebleumethodautomaticevaluation}.
Finally, image similarity analysis is conducted, comparing the generated charts with the provided ground truth chart images from the two datasets to gauge the accuracy of the data visualizations created by the agent.  Image similarity was determined with both the Structural Similarity Index Measure (SSIM) as well as multi-modal LLM as a judge~\cite{10.1007/978-3-031-70533-5_26,10.1007/978-3-031-88720-8_22}. SSIM measures structural differences between two images, ranging from 0 to 1, with 1 being perfect similarity. For the MM-LLM LLM-as-a-judge analysis, in-context (few-shot) prompting was used to compare the generated charts with the ground truth, with GPT4o-mini generating a perceptual quality score ranging from 0 to 100 for each chart when compared to the original~\cite{10.1007/978-3-031-72904-1_9}.  The appendix details the prompts used in the MM-LMM judge procedure.  The SSIM and perceptual quality scores were averaged for the baseline and agentic process results for all generated images. Thus, the agentic calculations have more data points than baseline calculations.

\vspace{1mm} \noindent \textbf{Text2Chart31 Results.} Table~\ref{tab:text2chart31-res} presents the results of our agentic process compared to the models published in the Text2Chart31 paper~\cite{pesaran-zadeh-etal-2024-text2chart31}. For the comparison, the models labeled SFT and SFT+RL are the supervised fine-tuning and reinforcement learning versions from the benchmark study. Our baseline results are similar to other closed-source models such as the GPT offerings and Claude3Opus, as well as the SFT and SFT+RL 8B models. However, our agentic runs showed major decreases in error rates in executing the code, dropping by more than half to two-thirds over the baseline and surpassing the results of the previously published models by far. The agentic few-shot run had an error rate of 4.5 percent, less than half of the SFT+RL 13B model at 9.2 percent for total error.  The agentic models also displayed strong performance across all chart types, with the weakest result being 13.2 percent error for 3D and Volumetric charts.  Pairwise plots had an error rate of only 1.5 percent. 

However, the metrics involving code similarity did not show similar improvements.  While comparisons to the published results might suffer from inconsistent alignment with their methodologies, the fact that our agentic scores are no better than the baseline indicates that code similarity might be of lesser importance, given the decreased error rate, especially with the image quality findings discussed below. More specifically, code similarity is not as important if the code does not run. Yet, we still show better performance than stronger models (GPT-4o) for METEOR. We do note that prior research results indicate that code similarity may not be a good indicator of visualization quality~\cite{10670425}.  Furthermore, the agentic process lends itself to creative GenAI problem-solving, which, in turn, produces deviations from the original code.

\vspace{1mm} \noindent \textbf{ChartX Results.} Table~\ref{tab:chartx-res} lists the results using the ChartX dataset. Note that the ChartX dataset was not originally used for text-to-chart generation~\cite{xia2025chartxchartvlmversatile}, hence we only compare to our baselines. As evidenced in the first dataset performance, the agentic methods greatly improve the successful execution of the generated code to produce the charts.  Error rates decline by approximately half to two-thirds.  Interestingly, while the few-shot approach improved the error rate for the baseline as expected, the result was reversed once the agentic process corrected the code.  Also interesting is the fact that the final zero-shot agentic results were almost identical between both the Text2Chart31 and this ChartX dataset at 4.5 and 4.6 percent, respectively.

\begin{figure}[t]
    \centering
    \includegraphics[width=.7\linewidth]{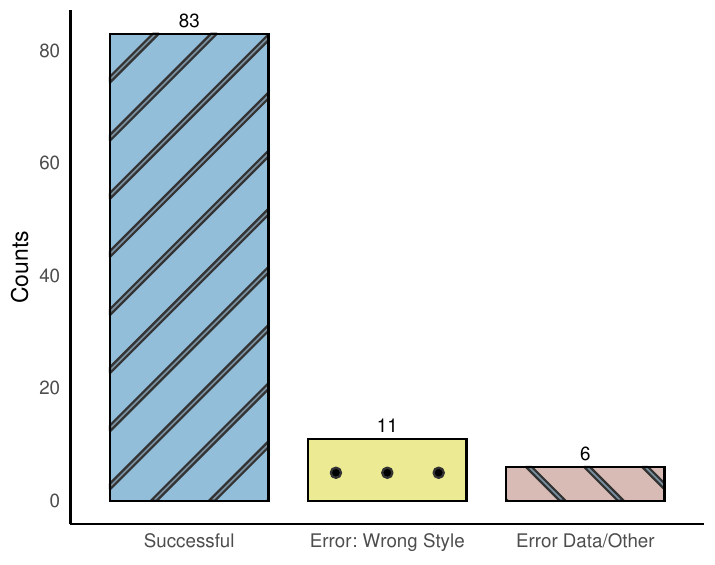}
    \caption{{Iteration Human Study Review (n=100)}}\vspace{-1em}
    \label{fig:man-error}
\end{figure}

\begin{figure}[t]
    \centering
    \includegraphics[width=.65\linewidth]{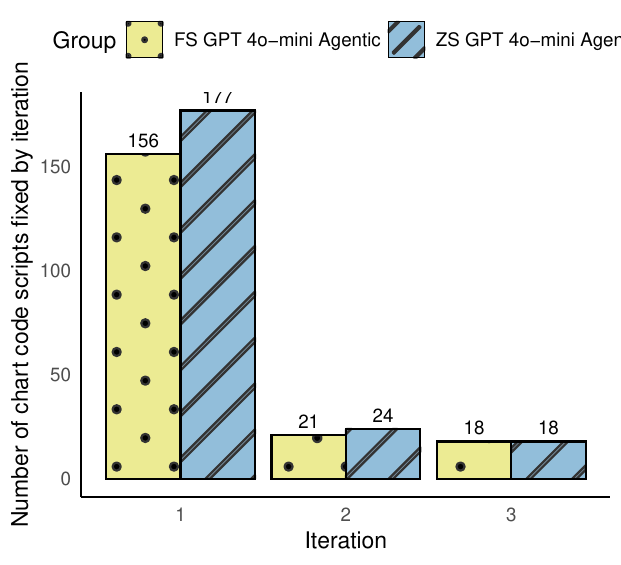}
    \caption{{Iteration Analysis for Text2Chart31.}}
    \label{fig:t2c31-iter}
\end{figure}

\begin{table}[t]
  \centering
  \resizebox{.98\linewidth}{!}{
  \begin{tabular}{lrrr}
    \toprule
    \textbf{Model} & \textbf{SSIM} & \textbf{MM-LLM as a Judge} \\
    \midrule
    \multicolumn{3}{c}{\textbf{Text2Chart31}} \\ \midrule
     GPT 4o-mini Agentic & 0.670 & 67.4 \\
     GPT 4o-mini Baseline & 0.672 & 66.7 \\ \midrule
    \multicolumn{3}{c}{\textbf{ChartX}} \\ \midrule
     GPT 4o-mini Agentic & 0.722 & 76.0 \\
     GPT 4o-mini Baseline & 0.722 & 76.4 \\\bottomrule
  \end{tabular}}
  \caption{Image Quality Analysis results on FS models.}\vspace{-1em}
  \label{tab:img-qual}
\end{table}

\begin{figure}[t]
    \centering
    \includegraphics[width=.7\linewidth]{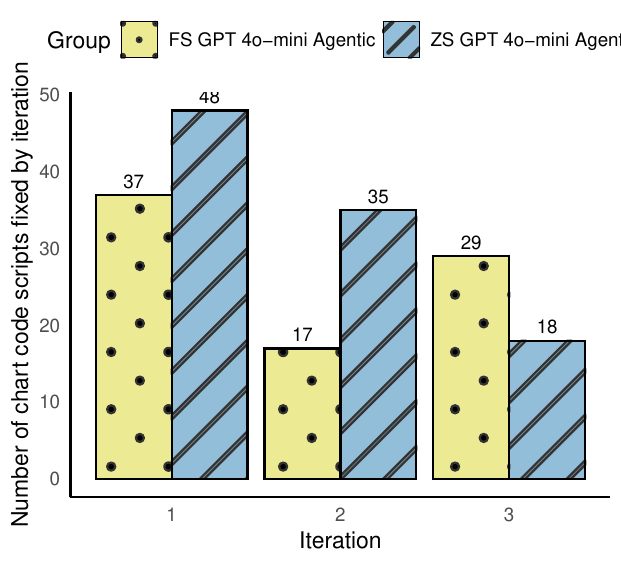}
    \caption{{Iteration Analysis for the ChartX dataset}}
    \label{fig:charx-iter}
\end{figure}

\begin{table}[t ]
  \centering
  \resizebox{\linewidth}{!}{
  \begin{tabular}{lrrr}
    \toprule
    \textbf{Model/Error Message} & \textbf{Count}  \\
    \midrule
         \multicolumn{2}{c}{\textbf{Text2Chart31}}  \\  \midrule
     stem() got an unexpected keyword argument "use\_line\_collection"  & 49 \\     
     Argument Z must be 2-dimensional  & 19 \\    
     name "np" is not defined  & 17  \\
    \midrule 
    \multicolumn{2}{c}{\textbf{ChartX}}  \\ \midrule
    No module named "mplfinance" & 39 \\
    No module named "squarify" & 25 \\
    All arrays must be of the same length & 18 \\
     \bottomrule
  \end{tabular}}
  \caption{Top 3 common Python interpreter errors for Iteration 1 on each dataset}\vspace{-1em}
  \label{tab:compile-errors}
\end{table}

\begin{figure*}[t]
\centering
\begin{subfigure}{0.49\textwidth}
    \centering
    \includegraphics[width=.8\linewidth]{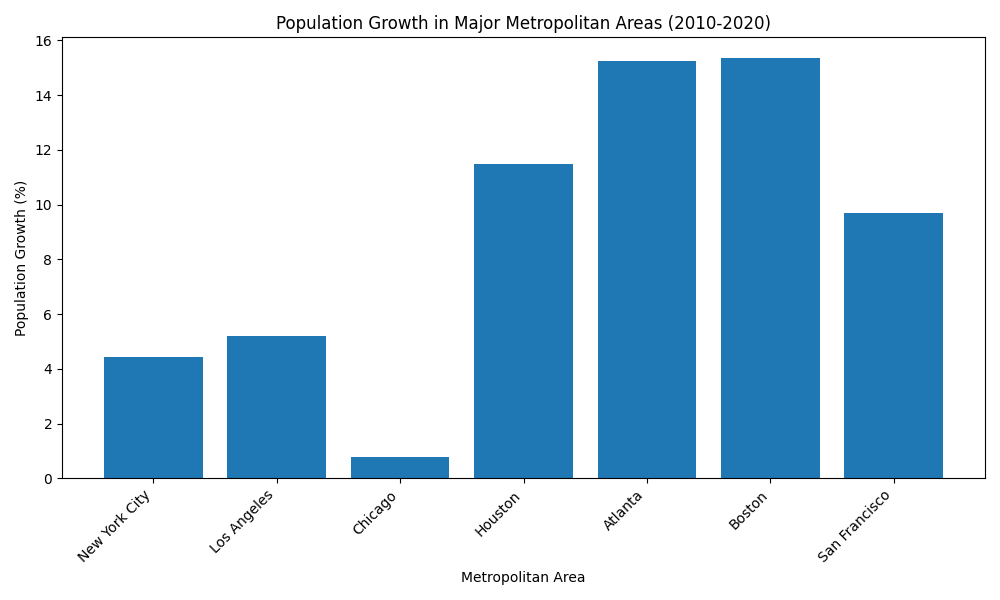}
    \caption{Ground Truth}
\end{subfigure}
\hfill
\begin{subfigure}{0.49\textwidth}
    \centering
    \includegraphics[width=.8\linewidth]{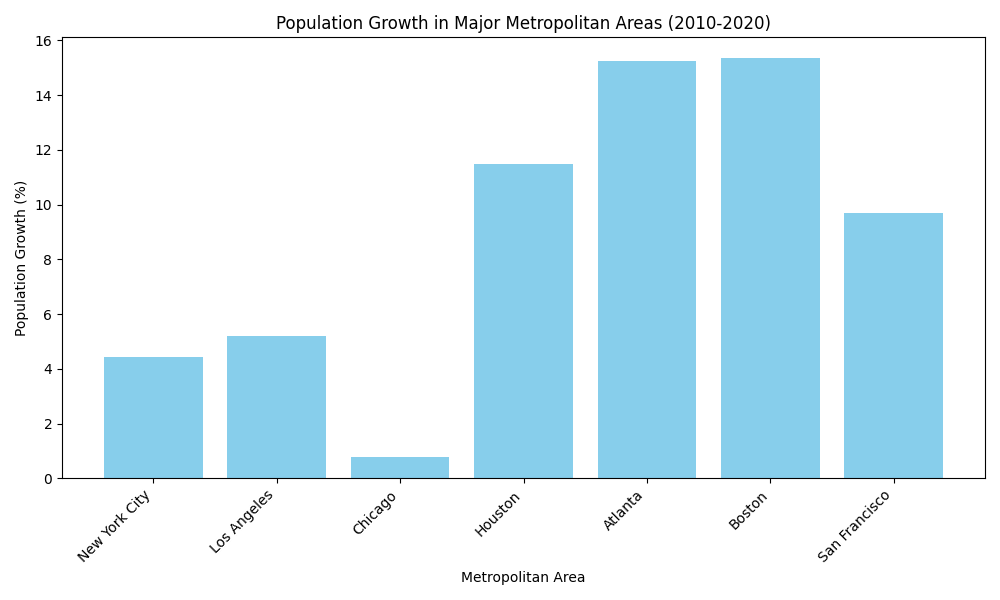}
    \caption{Agent Generated}
\end{subfigure}

\vfill 

\begin{subfigure}{0.49\textwidth}
    \centering
    \includegraphics[width=.8\linewidth]{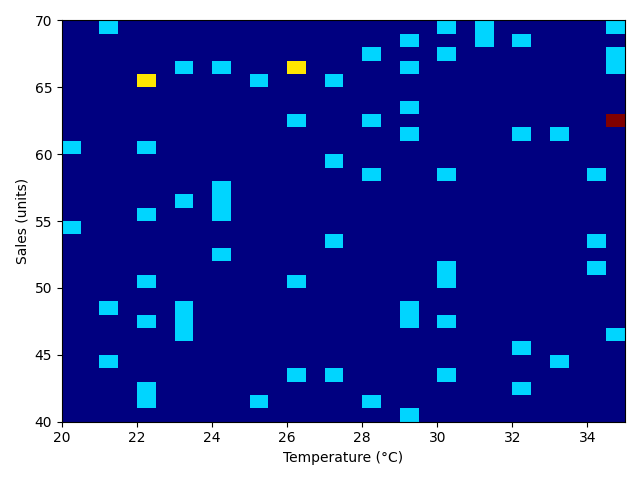}
        \caption{Ground Truth}
\end{subfigure}
\hfill
\begin{subfigure}{0.49\textwidth}
    \centering
    \includegraphics[width=.8\linewidth]{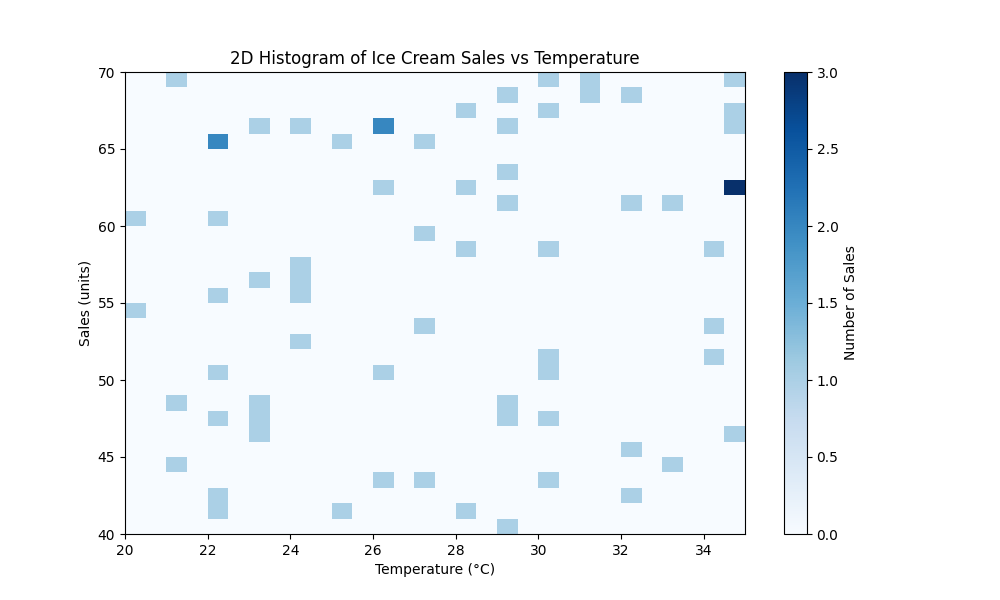}
    \caption{Agent Generated}
\end{subfigure}

\caption{Examples of visual similarity and variation in generated charts compared to ground truth. (a) and (b) show a bar chart comparing metropolitan population growth, with the agent-generated chart (b) closely matching the structure and values of the ground truth (a). (c) and (d) depict a 2D histogram of ice cream sales vs. temperature, where the agent-generated chart (d) captures the overall distribution but varies in visual encoding and bin density relative to the ground truth (c).}\vspace{-1em}
\label{fig:generations}
\end{figure*}

\vspace{1mm} \noindent \textbf{Ablation and Discussion.}
Overall, we perform a comprehensive ablation and error analysis to understand the current state of text-to-chart generation using an agent-based process. Specifically, we evaluate the impact of the number of iterations, how well accessible the charts are (about colorblindness), the general image quality compared to the ground-truth results, and a manual error analysis.

\vspace{1mm} \noindent \textbf{\textit{Iteration Ablation.}} To evaluate the impact of iterative repair, we conducted ablation studies examining how many chart scripts were successfully fixed at each step of the agentic pipeline. Figures~\ref{fig:t2c31-iter} and~\ref{fig:charx-iter} show the number of scripts corrected during the first, second, and third iterations for the Text2Chart31 and ChartX datasets, respectively. As expected, the few-shot (FS) GPT-4o-mini agent started with fewer initial code failures than the zero-shot (ZS) variant. Nevertheless, both settings benefited from additional repair attempts, particularly in the first two iterations. The largest gains occurred during the first pass, with diminishing but meaningful returns in subsequent iterations. These results highlight the value of multi-step reasoning and self-correction within agentic systems.

\vspace{1mm} \noindent \textbf{\textit{Color Blindness.}} Overall, we find that modern evaluation frameworks do not measure performance across many attributes needed to understand chart quality. Hence, we use an LLM-as-a-judge to evaluate how many generated charts pass standard colorblindness criteria (see the Appendix for the criteria and prompt). We manually evaluated the quality of the annotations on a random sample of 50 charts, and 46 were correctly labeled. For the Text2Chart31 dataset, we find that only 33.3\% were appropriate for color blindness. For the ChartX dataset, only 7.2\% of the generated charts were appropriate. This suggests that even though we can generate charts with a minimal error rate, substantial research is needed to make accessible charts.

\vspace{1mm} \noindent \textbf{\textit{Image Quality Analysis.}} Table~\ref{tab:img-qual} presents image quality scores for the agentic and baseline models across both datasets. Structural Similarity Index Measure (SSIM) values are nearly identical between the agentic and baseline models, indicating that the improved execution rates seen in Tables~\ref{tab:chartx-res} and~\ref{tab:text2chart31-res} do not come at the cost of chart quality. In other words, the agent generates more executable charts while maintaining visual similarity to the ground truth.

Similarly, perceptual judgments by a multimodal LLM (MM-LLM) show comparable scores across methods (ranging from 67 to 76 on a 100-point scale), further confirming that the agentic process does not degrade the perceived quality of the charts while resulting in much better execution performance. We do note that the prompts provided to both models contained only high-level descriptions (e.g., chart type and data values), without detailed formatting instructions, so some stylistic deviation from ground truth is expected.

Figure~\ref{fig:generations} illustrates close matches and common variations in chart outputs. Panels (a) and (b) depict a bar chart with minimal deviation: the agent-generated chart closely mirrors the structure and data of the ground truth, differing only in bar shading. In contrast, panels (c) and (d) show a 2D histogram with more noticeable differences. While the agent chart preserves the overall distribution, it employs a different colormap and includes a legend and title not present in the ground truth. These examples demonstrate that even when stylistic variations occur, the generated visualizations are plausible, faithful to the underlying data, and often enhanced with additional chart elements

\vspace{1mm} \noindent \textbf{\textit{Error Analysis.}} To better understand the nature of remaining errors, we manually reviewed 100 randomly sampled charts generated by our few-shot agentic system. We provide a summary of the analysis in Figure~\ref{fig:man-error}. Each chart was labeled as either \emph{Successful} or assigned one of two error categories: \emph{Wrong Style} (e.g., incorrect chart type, missing stylistic elements) or \emph{Error Data/Other} (e.g., incorrect values, axis misalignment, or malformed outputs not caught by execution checks). As shown in Figure~\ref{fig:man-error}, 83\% of the charts were deemed fully successful. Among the 17 failures, 11 involved stylistic mismatches that still plausibly conveyed the intended message (e.g., a line chart instead of a bar chart). At the same time, only 6 were due to data-related or structural errors. This analysis suggests that while execution correctness has largely been addressed, future improvements should focus on semantic and stylistic fidelity.

To better understand the kinds of issues the agent addresses, we analyzed the most frequent runtime errors encountered during the first iteration of the repair process. As shown in Table~\ref{tab:compile-errors}, the top errors for the Text2Chart31 dataset included incorrect keyword arguments (e.g., unsupported parameters like use\_line\_collection), shape mismatches (e.g., expecting a 2-dimensional input), and missing or incorrectly referenced libraries (e.g., undefined variables like np). For ChartX, the most common errors involved missing third-party libraries (mplfinance, squarify) and array length mismatches. The agent typically responds to the first error surfaced by the Python runtime, treating the repair process as a sequential correction task. As a result, deeper or secondary issues are often uncovered in subsequent iterations, supporting the design choice of a multi-round repair loop.

\vspace{1mm} \noindent \textbf{\textit{Implications}.}
Our findings suggest that text-to-chart generation, as currently defined by benchmarks like \textsc{Text2Chart31} and \textsc{ChartX}, is approaching a performance ceiling concerning execution correctness. By introducing a lightweight multi-agent framework that separates drafting, execution, and iterative repair, we reduced error rates by over 5 percentage points without relying on supervised fine-tuning or reinforcement learning. This indicates that when paired with execution-aware self-correction, structured prompting alone can achieve or exceed the performance of more computationally expensive methods.

However, execution success alone does not guarantee semantic accuracy, visual clarity, or accessibility. While our system maintains visual quality (via SSIM and multimodal LLM judgment), manual analysis shows that some outputs still deviate stylistically or semantically from the intended chart. Additionally, only 7-33\% of generated charts meet basic color accessibility standards, underscoring a critical gap in current evaluation protocols.

These results call for reorienting future work in this space, from reducing execution failure to enhancing chart readability, aesthetics, and inclusivity. Benchmarks must evolve to reflect real-world use cases where visual clarity and usability are as important as syntactic correctness. Agentic systems, by enabling structured reasoning and self-verification, offer a promising foundation for tackling these higher-level challenges.

\section{Conclusion}
We presented a lightweight agentic framework that significantly improves text-to-chart generation by reducing execution errors while preserving image quality, all without fine-tuning or reinforcement learning. Our analysis shows that single-prompt LLMs (whether zero- or few-shot) still fail frequently, whereas multi-agent repair loops offer more robust performance at low cost.

Beyond execution, we highlight the need to shift focus toward semantic fidelity, visual clarity, and accessibility. Paired with image quality analysis, agent-based approaches can help address these broader challenges. Future work should explore how these methods can enhance real-world applications, especially for users with visual or cognitive impairments~\cite{10.1145/3654777.3676414,10.1145/3663548.3675660}.

\section*{Acknowledgments}
This material is based upon work supported by the National Science Foundation (NSF) under Grant~No. 2145357.

\section*{Limitations}
While our proposed framework demonstrates strong performance in generating LLM-based data visualizations using a multi-agent approach, several limitations remain. First, our evaluation focuses on two benchmark datasets, Text2Chart31 and ChartX. Both datasets are synthetically generated and may not reflect the full complexity of real-world chart generation. Future work should consider additional datasets that include human-authored chart descriptions and real-world data to better evaluate generalizability. Second, the system uses a single proprietary model, GPT-4o-mini, for drafting and repair. Exploring open-source LLMs as alternatives could improve transparency and reproducibility. Testing a broader set of LLM providers may also reveal model-specific strengths or weaknesses. Finally, our agentic framework uses a simple two-agent setup without memory, planning, or retrieval modules. More advanced agent designs supporting longer-term reasoning or richer tool use could improve performance, especially in more complex or multi-step charting scenarios. Finally, we note that we use AI to help improve our writing of this manuscript.

\bibliography{custom}

\appendix

\section{Appendix}
\label{sec:appendix}

\noindent \textbf{Prompts for Baseline}
The prompts used to create the baseline charts are derived from the code in the Text2Chart31 dataset~\cite{pesaran-zadeh-etal-2024-text2chart31}.  These were modified for the few-shot run by including two example descriptions and generated code from the training set from that data source.

\begin{center}
\vspace{-.25em}
\tcbset{
    colframe=black,
    colback=white,
    boxrule=0.5mm,
    arc=3mm,
    width=1\linewidth,
    boxsep=5pt,
    left=5pt,
    right=5pt,
    top=5pt,
    bottom=5pt
}
\begin{tcolorbox}[colback=gray!5!white, colframe=black, fontupper=\small, width=\linewidth, boxsep=4pt, title=System Prompt]
\textit{You are good at generating complete python code from the given chart description.}
\end{tcolorbox}
\vspace{-.25em}
\end{center}

\begin{center}
\vspace{-.25em}
\tcbset{
    colframe=black,
    colback=white,
    boxrule=0.5mm,
    arc=3mm,
    width=1\linewidth,
    boxsep=5pt,
    left=5pt,
    right=5pt,
    top=5pt,
    bottom=5pt
}
\begin{tcolorbox}[colback=gray!5!white, colframe=black, fontupper=\small, width=\linewidth, boxsep=4pt, title=User Instructions]
\textit{Your task is to generate a complete Python code for the given description. Make sure to include all necessary libraries.}

\texttt{Description: Description\_text}

\textit{Please generate the corresponding code that generates the plot that has the above description.}
\texttt{Code:}
\begin{verbatim}
```import matplotlib.pyplot as plt
import pandas as pd
import numpy as np
\end{verbatim}
\end{tcolorbox}
\vspace{-.25em}
\end{center}

\noindent \textbf{Prompts for Agentic Process} 
The prompts used in the agentic process are presented below. The LangChain framework uses multiple tools to complete the agentic tasks, thus the prompts are embedded within each tool.

\begin{center}
\vspace{-.25em}
\tcbset{
    colframe=black,
    colback=white,
    boxrule=0.5mm,
    arc=3mm,
    width=1\linewidth,
    boxsep=5pt,
    left=5pt,
    right=5pt,
    top=5pt,
    bottom=5pt
}
\begin{tcolorbox}[colback=gray!5!white, colframe=black, fontupper=\small, width=\linewidth, boxsep=4pt, title=Reflection Tool]
\textit{The following Python code produced an error:}

\texttt{code}

\texttt{Error: {error}}

\textit{Identify the root cause of the error. Provide a suggestion to fix ONLY the problematic lines,
    explicitly specifying which parts of the original code should REMAIN UNCHANGED. Return the complete code with the suggested modifications inserted.}
\end{tcolorbox}
\vspace{-.25em}
\end{center}

\begin{center}
\vspace{-.25em}
\tcbset{
    colframe=black,
    colback=white,
    boxrule=0.5mm,
    arc=3mm,
    width=1\linewidth,
    boxsep=5pt,
    left=5pt,
    right=5pt,
    top=5pt,
    bottom=5pt
}
\begin{tcolorbox}[colback=gray!5!white, colframe=black, fontupper=\small, width=\linewidth, boxsep=4pt, title=Rewriter Tool]
\textit{system\_message = 
        You are an expert Python code rewriter. Your task is to rewrite Python code based strictly on the user's suggestions.
        - DO NOT modify any part of the code that is not explicitly mentioned in the suggestion.
        - Ensure that the rewritten code is functional, error-free, and adheres to Python syntax rules (e.g., indentation, brackets, braces).
        - Return ONLY the complete revised code without explanations, comments, or Markdown formatting.
        - Follow instructions EXACTLY as provided."""
    }

    \textit{user\_message = 
        """Rewrite the following Python code based on this suggestion:}

        \texttt{Original Code:
        {code}}

        \texttt{Suggestion:
        {suggestion}
    }
\end{tcolorbox}
\vspace{-.25em}
\end{center}

\noindent \textbf{Prompts for MM-LLM as a judge} For the MM-LLM as a judge analysis, GPT4o-mini was prompted to rate the quality of each of the baseline and agentic charts which were generated.
\begin{center}
\vspace{-.25em}
\tcbset{
    colframe=black,
    colback=white,
    boxrule=0.5mm,
    arc=3mm,
    width=1\linewidth,
    boxsep=5pt,
    left=5pt,
    right=5pt,
    top=5pt,
    bottom=5pt
}
\begin{tcolorbox}[colback=gray!5!white, colframe=black, fontupper=\small, width=\linewidth, boxsep=4pt, title=MM-LLM Prompt]
\textit{system\_message = 
        For two shown images, the human perceptual quality score of the first image is 50.}
    \textit{Based on this example, assign a perceptual quality score to the second image in terms of perceptual similarity.}
    \textit{The score must range from 0 to 100, with a higher score denoting better image quality.}
    \textit{Return the result as a JSON object in this format: {\"score\": <integer>}.}

    \textit{user\_message = }
    
    \textit{\{"type": "image\_url"\}}
    
    \textit{"image\_url":} 
    
    \textit{{"url": f"data:image/png;base64,{base64\_image1}}}
     
    \textit{\{"type": "image\_url",\}}
    
    \textit{"image\_url":}
    
    \textit{{     "url": f"data:image/png;base64,{base64\_image2}"}}
                          
\end{tcolorbox}
\vspace{-.25em}
\end{center}

\noindent \textbf{Color Blindness Analysis} For the color blindness analysis, GPT4o-mini was prompted to evaluate whether the generated chart was appropriate or not based on color blindness criteria.  To assess whether generated charts are appropriate for color-vision-deficient users, we created an LLM-based judgment prompt grounded in accessibility research. Specifically, we drew on practical recommendations from \citet{muth2020colorblind}, who highlights common failure modes in data visualizations for colorblind readers, such as over-reliance on hue and the use of indistinguishable color pairs like red/green, purple/blue, or pink/grey when luminance is held constant. These combinations, while legible to users with typical vision, can render visual encodings ambiguous or unreadable for individuals with color deficiencies.

Our system prompt was designed to reflect these insights, emphasizing three critical criteria: (1) redundant encoding, such as pairing color with shape, text, or iconography; (2) contrast in lightness, which improves discernibility even when hue perception is impaired; and (3) legend readability, including whether text labels and gradients remain interpretable under color vision simulation.

\begin{center}
\vspace{-.25em}
\tcbset{
    colframe=black,
    colback=white,
    boxrule=0.5mm,
    arc=3mm,
    width=1\linewidth,
    boxsep=5pt,
    left=5pt,
    right=5pt,
    top=5pt,
    bottom=5pt
}
\begin{tcolorbox}[colback=gray!5!white, colframe=black, fontupper=\small, width=\linewidth, boxsep=4pt, title=Color Blindness Prompt]
\textit{system\_message = 
        You will be given a data-visualization (image). Decide whether the visualization is appropriate for viewers with color-vision deficiencies.}
\textit{How to judge: }
\textit{1. Do not rely on hue alone. Look for additional cues such as shapes, icons, text labels, patterns, or distinct light-dark contrasts.}
\textit{2. Avoid problem pairs. Red + green, red + brown, green + brown, purple + blue, and pink + turquoise of similar lightness are hard to tell apart.}
\textit{3. Prefer safe palettes. Blue vs. orange/red, or any two colors that differ clearly in lightness, usually work well.}
\textit{Check gradients. Gradients should vary in lightness, not just hue.}
\textit{Overall clarity. Annotations, legends, and labels must still be readable when colors are altered by common forms of color-blindness.}
\textit{Output format: Just return \"Appropriate\" or \"Not appropriate\", do NOT return anything else.}
\textit{                        "Return the result as a JSON object in this format: {\"Judgment\": <string>}.}

\textit{user\_message = }
    
    \textit{{"type": "image\_url",}}
    
    \textit{"image\_url":}
    
    \textit{{     "url": f"data:image/png;base64,{base64\_image2}"}}

\end{tcolorbox}
\vspace{-.25em}
\end{center}
\noindent \textbf{Error Analysis} Examples of various error either in the graph generation or the color blindness assessment.

\begin{figure}[htbp]
    \centering
    \begin{subfigure}[t]{0.45\textwidth}
        \centering
        \includegraphics[width=\textwidth]{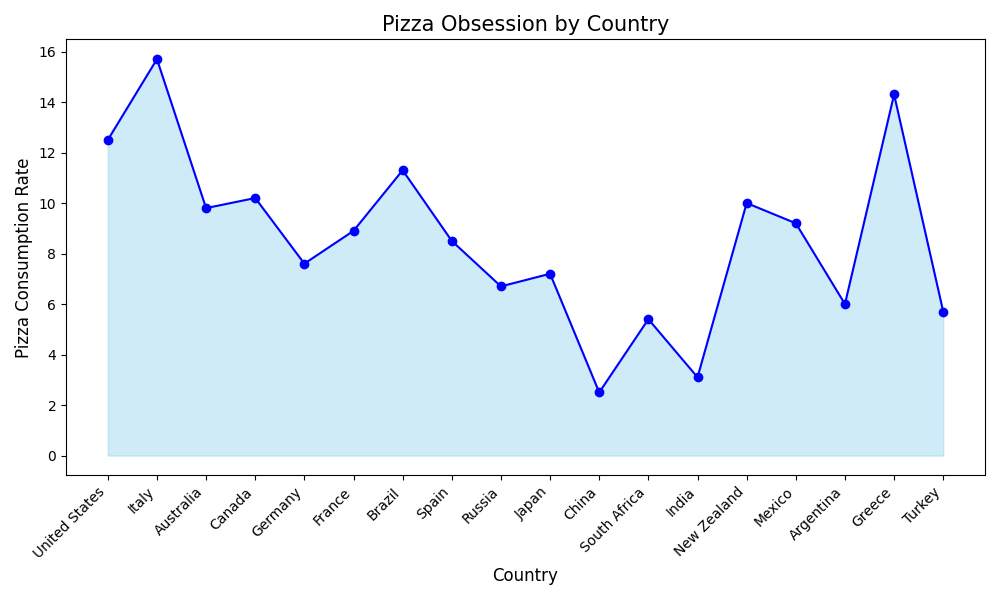}
        \caption{Ground Truth}
        \label{fig:sub1}
    \end{subfigure}
    \hfill
    \begin{subfigure}[t]{0.45\textwidth}
        \centering
        \includegraphics[width=\textwidth]{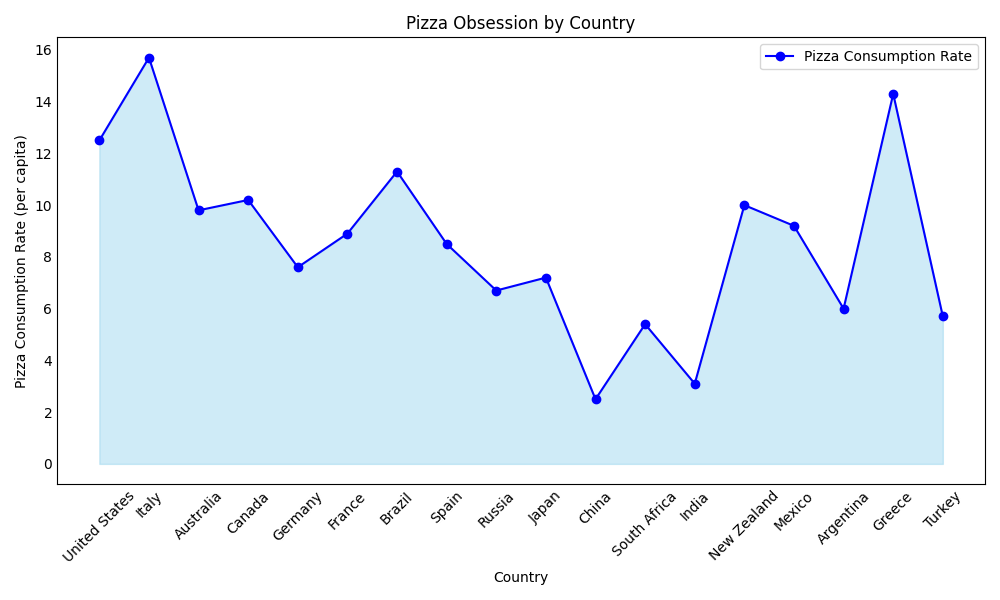}
        \caption{Agentic}
        \label{fig:sub2}
    \end{subfigure}
    \caption{Example of the agent generating a successful chart}
    \label{fig:success}
\end{figure}

\begin{figure}[htbp]
    \centering
    \begin{subfigure}[t]{0.45\textwidth}
        \centering        \includegraphics[width=\textwidth]{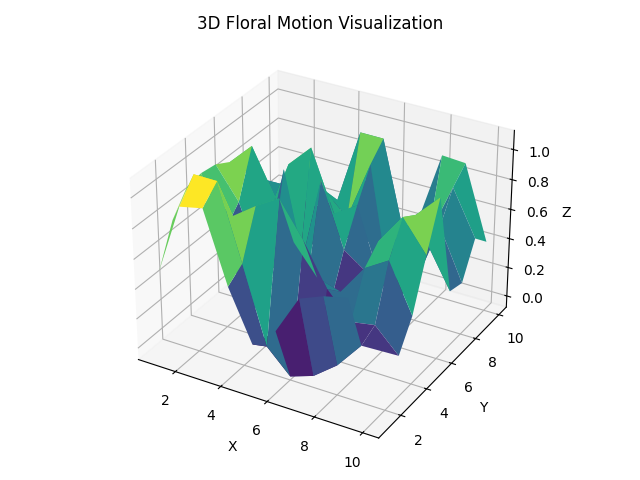}
        \caption{Ground Truth}
        \label{fig:sub1}
    \end{subfigure}
    \hfill
    \begin{subfigure}[t]{0.45\textwidth}
        \centering        \includegraphics[width=\textwidth]{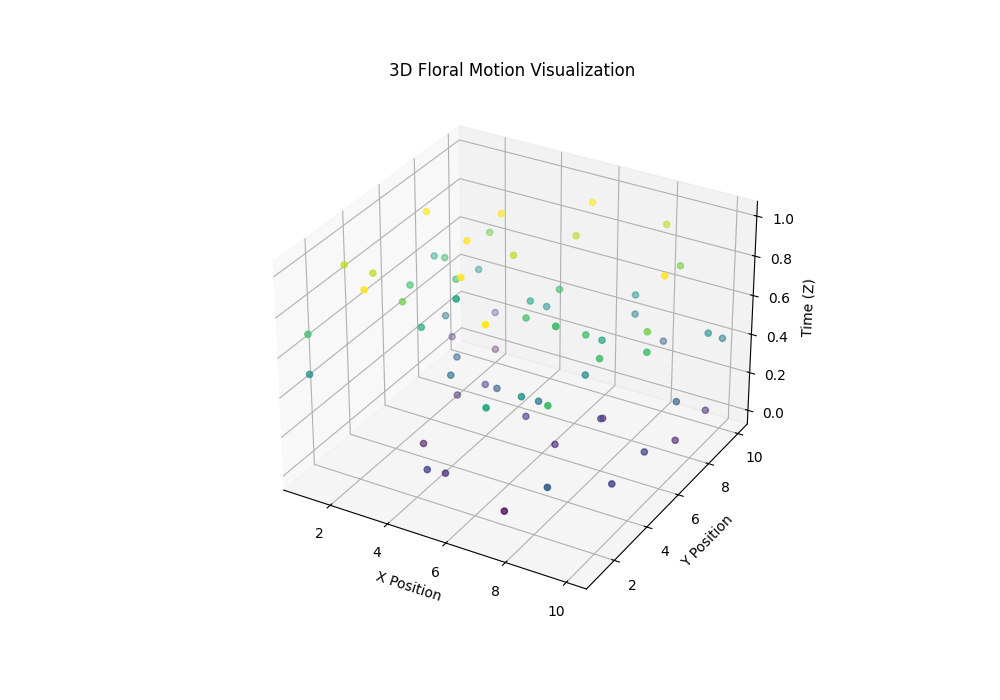}
        \caption{Agentic}
        \label{fig:sub2}
    \end{subfigure}
    \caption{Example of the agent generating the wrong style chart}
    \label{fig:wrong style}
\end{figure}

\begin{figure}[htbp]
    \centering
    \begin{subfigure}[t]{0.45\textwidth}
        \centering        \includegraphics[width=\textwidth]{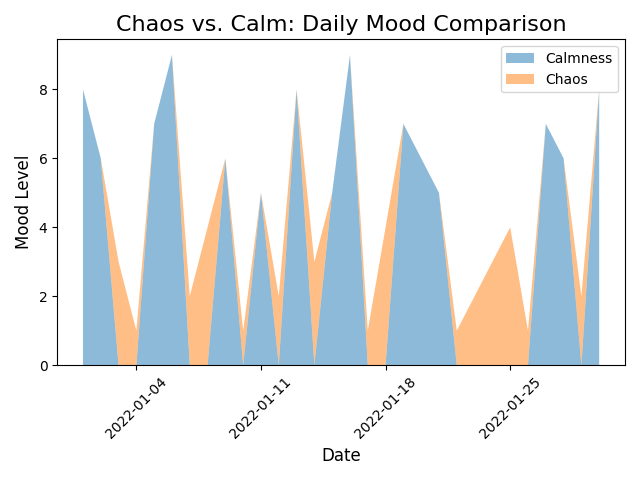}
        \caption{Ground Truth}
        \label{fig:sub1}
    \end{subfigure}
    \hfill
    \begin{subfigure}[t]{0.45\textwidth}
        \centering        \includegraphics[width=\textwidth]{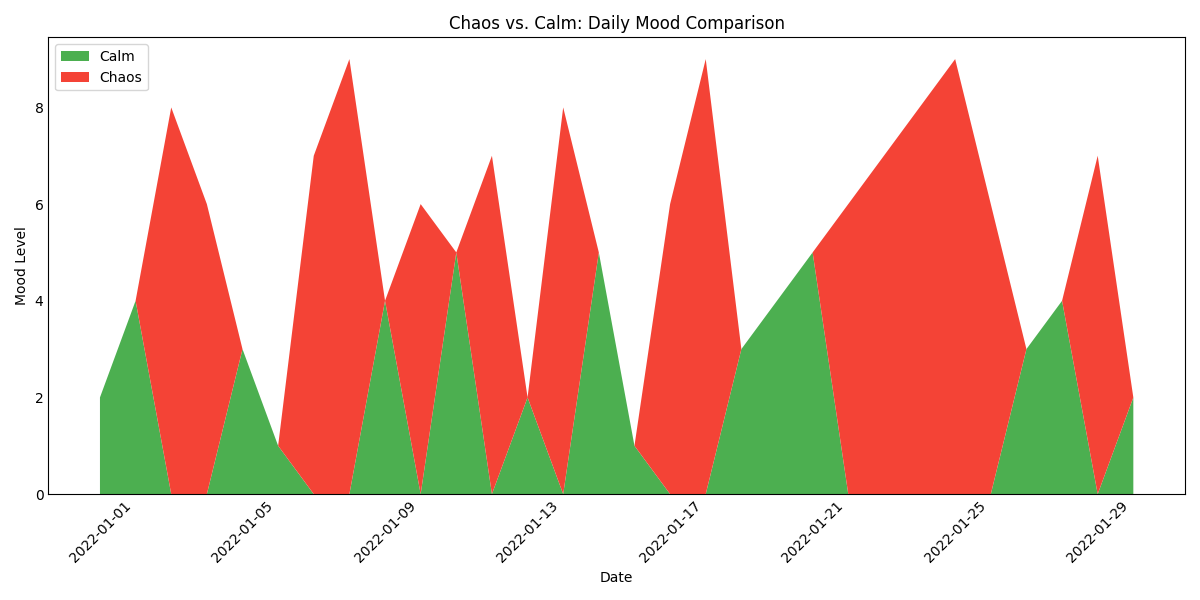}
        \caption{Agentic}
        \label{fig:sub2}
    \end{subfigure}
    \caption{Example of the agent generating the wrong data for a chart}
    \label{fig:wrong data}
\end{figure}

Figure~\ref{fig:success} displays an example of the agentic process generating a successful chart. Figure~\ref{fig:wrong style} displays an example of the agentic process generating the wrong style of chart. Figure~\ref{fig:wrong data} displays an example of the agentic process generating the  chart with incorrect data, as the labels are interchanged. 

\begin{figure}[t]
    \centering    \includegraphics[width=.7\linewidth]{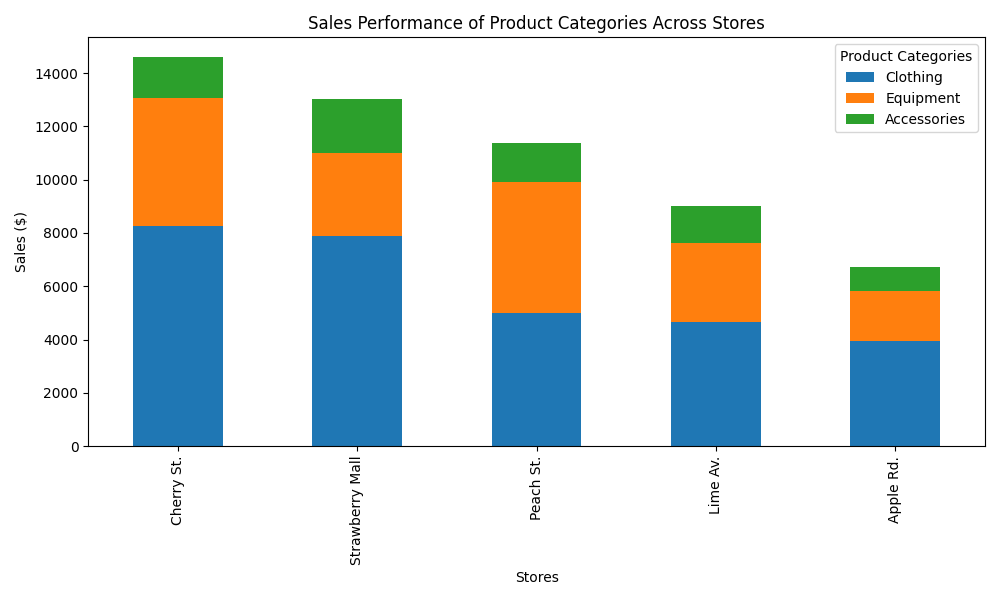}
    \caption{{Color Blindness assessment correct}}
    \label{fig:color blindness correct}
\end{figure}

\begin{figure}[t]
    \centering    \includegraphics[width=.7\linewidth]{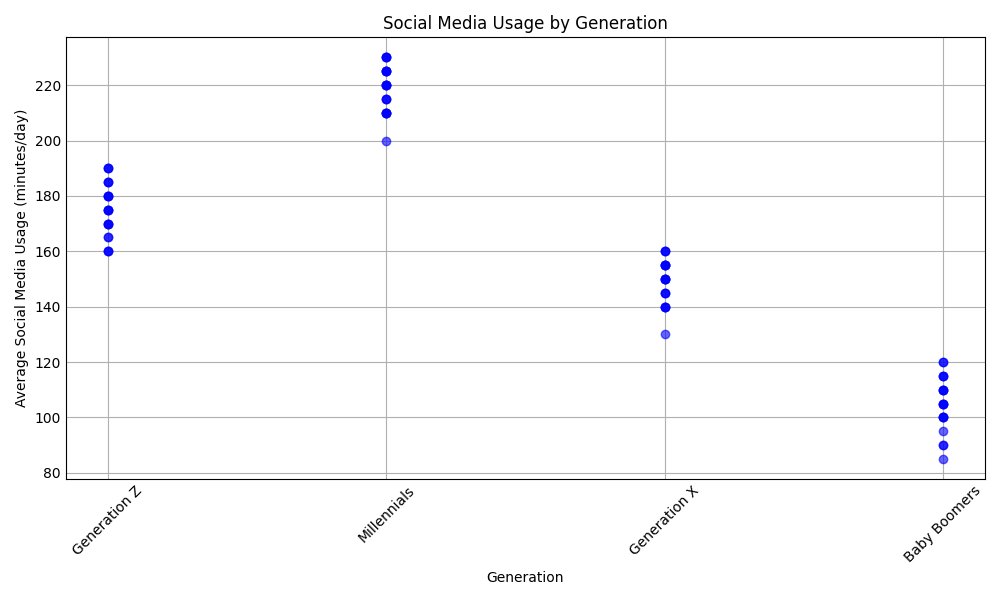}
    \caption{{Color Blindness assessment error}}
    \label{fig:color blindness example}
\end{figure}

For the color blindness analysis, Figure~\ref{fig:color blindness correct} displays an example of GPT4o-mini correctly assessing a chart as being "Not Appropriate" for color blindness. The chart includes both blue and green, which may cause issues for viewers with visual impairments. 
In contrast, Figure~\ref{fig:color blindness example} displays an example of GPT4o-mini incorrectly assessing a chart as being "Not Appropriate" for color blindness.  The chart only uses one shade of blue along with the grey border and gridlines, so this is less likely to cause issues for the visually impaired.

\end{document}